\documentclass[final]{cvpr}

\usepackage{times}
\usepackage{epsfig}
\usepackage{graphicx}
\usepackage{amsmath,amssymb,mathrsfs}
\usepackage{amsthm}
\usepackage{booktabs}
\usepackage{tabularx}
\usepackage{comment}
\usepackage{makecell}
\usepackage{cite}
\usepackage{mhsetup}
\usepackage{mathtools}
\usepackage{multirow}
\usepackage{float}

\usepackage[pagebackref=true,breaklinks=true,colorlinks,bookmarks=false]{hyperref}

\begin{document}

%%%%%%%%% TITLE
\title{Multi-Modal Temporal Convolutional Network for Anticipating Actions in Egocentric Videos}

\author{Olga Zatsarynna, Yazan Abu Farha and Juergen Gall\\
University of Bonn \\
Germany\\
{\tt\small \{zatsarynna, abufarha, gall\}@iai.uni-bonn.de}
\\}

\maketitle

%%%%%%%%% ABSTRACT
\begin{abstract}
Anticipating human actions is an important task that needs to be addressed for the development of reliable intelligent agents, such as self-driving cars or robot assistants. While the ability to make future predictions with high accuracy is crucial for designing the anticipation approaches, the speed at which the inference is performed is not less important. Methods that are accurate but not sufficiently fast would introduce a high latency into the decision process. Thus, this will increase the reaction time of the underlying system. This poses a problem for domains such as autonomous driving, where the reaction time is crucial. In this work, we propose a simple and effective multi-modal architecture based on temporal convolutions. Our approach stacks a hierarchy of temporal convolutional layers and does not rely on recurrent layers to ensure a fast prediction. We further introduce a multi-modal fusion mechanism that captures the pairwise interactions between RGB, flow, and object modalities. Results on two large-scale datasets of egocentric videos, EPIC-Kitchens-55 and EPIC-Kitchens-100, show that our approach achieves comparable performance to the state-of-the-art approaches while being significantly faster.
\end{abstract}

%%%%%%%%% BODY TEXT
\section{Introduction}

Anticipating future events is of great importance for intelligent agents. There are many real-world scenarios, where apart from recognizing what is happening in the current moment, one also needs to make predictions about the future. For example, autonomous driving systems need to anticipate pedestrians movement to avoid collisions. Another field of application is assistive robotics, where the ability of robots to anticipate future human activities allows for smoother and more productive interactions. In our work, we focus on human activity anticipation since it is a challenging yet crucial task for an intelligent system to be deemed as such.

\begin{figure}[]
    \centering
    \includegraphics[width=\columnwidth]{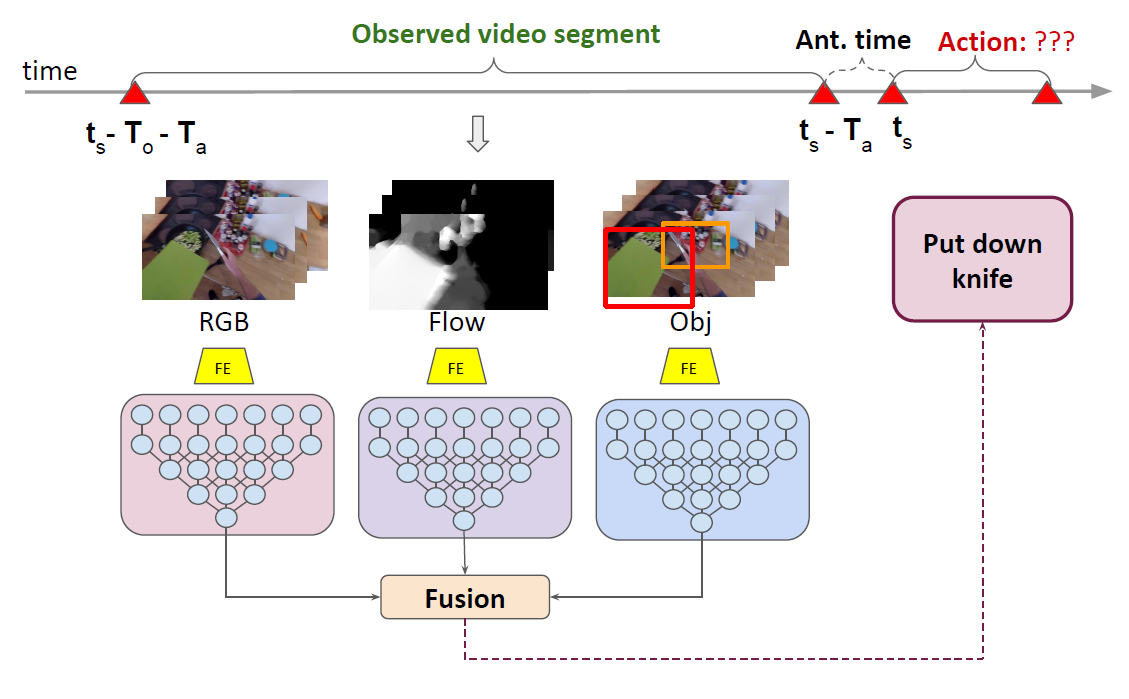}
    \caption{The short-term action anticipation task predicts the next unobserved action $T_a$ seconds before it occurs. To address this task, we propose a multi-modal approach based on temporal convolutional networks that achieves state-of-the-art results while being faster than traditional RNN-based approaches.}
    \label{fig:intro}
\end{figure}

In recent years, the number of works addressing the task of action anticipation has experienced a substantial increase. Generally, one could subdivide these works into two categories based on the time horizon of anticipation they tackle. While some works try to anticipate several actions into the future (long-term anticipation)~\cite{Farha_2018_CVPR, Ke_2019_CVPR, Gammulle2019ForecastingFA, farha2020gcpr}, others aim at anticipating only the next action at a fixed anticipation time based on the recent observations preceding it (short-term anticipation)~\cite{Vondrick_2016_CVPR, Gao_2017_BMVC, Jain_2016_ICRA, furnari2019rulstm}. In our work, we deal with the second setting as illustrated in Figure~\ref{fig:intro}. 

Initially, this task has been addressed by predicting representations of the future frames and anticipating the actions by training a classifier on them~\cite{Vondrick_2016_CVPR}. Such approaches, while being successful on videos shot from the third-person view, do not perform well on egocentric videos captured from the first-person view. 
%(shot from the first-person view, see Figure~\ref{fig:intro}), due to the challenges posed by a greater mobility of the camera, resulting in large motions and fast changes in appearance. 
Egocentric action anticipation has been addressed in the work of Furnari~\etal~\cite{furnari2019rulstm}, who introduced a multi-modal LSTM-based~\cite{hochreiter_1997_neuralcomp} encoder-decoder network.
%RU-LSTM - a three-branch network, with each branch consisting of two LSTM  modules - R-LSTM and U-LSTM, responsible for summarization of the past observations and for making predictions about the future respectively. 
Recently, several new methods have been proposed~\cite{sener-2020-temporal, Dessalene_2021, Liu_2020_ECCV, Wu_2021} that show improved performance in the egocentric action anticipation.

While these methods demonstrate better performance in the egocentric short-term action anticipation, little attention has been paid to the effectiveness of the training and inference procedures of the methods. Many of the above mentioned works, in particular~\cite{Dessalene_2021, furnari2019rulstm, Wu_2021}, use recurrent layers for performing temporal sequence modelling. However, in recent years it has been repeatedly indicated~\cite{bai2018empirical, gehring17a, Lea_2017_CVPR, Oord_16_wave} that for many sequence processing tasks convolution-based methods are much faster than canonical recurrent layers such as LSTMs, and still show a similar or even superior performance. Convolution-based architectures are also easier to train, since they do not possess notorious drawbacks of recurrent layers such as vanishing gradients and inability to model long-term dependencies.
On a different note, some of the action anticipation works~\cite{Liu_2020_ECCV, Dessalene_2021} also gather and incorporate additional data or annotations into training, which is a costly and labor intensive process.

In our work, we propose a model that addresses the previously mentioned limitations. We introduce a multi-modal network based on a hierarchy of temporal convolutions. %, inspired by \cite{Lea_2017_CVPR}.
Our network consists of three parallel branches where each branch operates on features extracted from RGB, optical flow or object modalities. To fuse these modalities, we introduce a multi-modal fusion mechanism that captures both mutual and pairwise interactions between the different branches. 
In contrast to previous approaches~\cite{Liu_2020_ECCV, Dessalene_2021}, our model 
does not require any additional data or annotations. 
%other than that made publicly available by egocentric datasets. 
We evaluate our approach on two large-scale datasets of egocentric videos: EPIC-Kitchens-55~\cite{Damen2018EPICKITCHENS, Damen2020Collection_epic_55} and EPIC-Kitchens-100~\cite{Damen2020RESCALING}.
We show that our model achieves comparable results to the state-of-the-art while being at least two times faster during both training and inference. 

%-------------------------------------------------------------------------

\section{Related Work}
\subsection{Action Anticipation in Videos}
There are several lines of work in the area of action anticipation that differ in the time horizon of predictions. Some approaches focus on long-term predictions. That is, given a subset of observed actions, they aim to predict multiple actions into the future or even all subsequent actions. In the work by Abu Farha~\etal~\cite{Farha_2018_CVPR}, two approaches for long-term anticipation have been proposed, based on RNN and CNN. The RNN-based method predicts labels and lengths of the upcoming actions by feeding its predicted values back to the network for future predictions. The CNN-based approach, unlike the previous one, predicts all future actions in a single step. It makes predictions by encoding both its input and output in a matrix form. To avoid intermediate computations and accumulation of errors, Ke~\etal~\cite{Ke_2019_CVPR} introduced a time-conditioned method that anticipates long-term actions in one shot. Gammulle~\etal~\cite{Gammulle2019ForecastingFA} proposed a network that models long-term relationships within the input sequence with the help of a neural memory module, whose refined output is then used to make action predictions.  In~\cite{farha2020gcpr}, a sequence-to-sequence model is used to predict future activities and their durations. Additionally, the authors leveraged cycle consistency over time by predicting past actions on the basis of the future predictions made by the network. In contrast to these approaches, we focus on short-term action anticipation from egocentric videos.

For the task of short-term action anticipation, the goal is to forecast an action several seconds prior to its occurrence. State-of-the-art methods usually take the most recent observations into account and predict actions up to several seconds into the future. For example, Vondrick~\etal~\cite{Vondrick_2016_CVPR} proposed a mixture of regression networks to learn a representation of a frame one second in the future based on the frame at the current time step. Then, to predict the action, they categorize the predicted representations with a classifier network. Gao~\etal~\cite{Gao_2017_BMVC} further extended the previous idea and introduced an encoder-decoder network that anticipates a sequence of future representations based on an observed sequence of representations, instead of just a single representation. In the work of Jain~\etal~\cite{Jain_2016_ICRA}, the authors also used an encoder-decoder network, albeit with several modalities and a loss that exponentially increases with time to prevent overfitting and encourage early anticipation. Different from previous works, Damen~\etal~\cite{Damen2018EPICKITCHENS} leveraged an action recognition network based on TSN \cite{TSN2016ECCV} for the task of action anticipation. During training, the network receives the observed segment preceding the action of interest as input, while the corresponding label is set to the category of the action that needs to be predicted. 
%Among the most recent approaches are \cite{furnari2019rulstm, sener-2020-temporal, Liu_2020_ECCV, Camporese2020KnowledgeDF, Miech_2019_CVPR_Workshops}.
Miech~\etal~\cite{Miech_2019_CVPR_Workshops} proposed to predict future actions by averaging predictions of two complementary modules: predictive and transitional, where the predictive model directly anticipates the upcoming action, and the transitional model is constrained to at first output the current action and then use the acquired information to anticipate the future. In~\cite{furnari2019rulstm}, the RU-LSTM network was introduced, that consists of two LSTM networks. The first LSTM summarizes the past observations whereas the second one predicts the future actions. Camporese~\etal~\cite{Camporese2020KnowledgeDF} further proposed to extend the RU-LSTM with label smoothing to mitigate over-confident predictions and make their system more uncertainty-aware. Sener~\etal~\cite{sener-2020-temporal} proposed a framework based on non-local blocks~\cite{wang_2018_CVPR}, that aggregates multi-scale features from the video by computing interaction between recent and distant observations. The resulting features are then used to anticipate both short-term and long-term actions. Liu~\etal~\cite{Liu_2020_ECCV} explicitly incorporate intentional hand movement as an anticipatory representation of actions. They jointly model and predict hand trajectories, interaction hotspots and labels of future actions.
Dessalene~\etal~\cite{Dessalene2020EgocentricOM} proposed to use a Graph Convolutional Network (GCN) to model long-term temporal semantic relations between actions based on contact information. They use the constructed graph representations along with appearance features to make anticipation about the future actions.
In contrast to these approaches, our approach relies on temporal convolutions to capture dependencies in the input sequence and predict the future action.

\subsection{Anticipation of other Modalities}
While in our work we address the anticipation of human activities, there are numerous efforts that address the anticipation of other modalities. Anticipation of human trajectories and motion is a popular task that has been addressed in many works~\cite{ziebart_2009_RSJ, kitani_2012_ECCV, fouhey_2014_CVPR, alahi_2016_CVPR}. Another line of work deals with predictions of future human poses~\cite{fragkiadaki_2015_ICCV, Jain_2016_CVPR, Martinez_2017_CVPR, hernandez_2019_ICCV}. Also, many approaches have been proposed for prediction of future semantic segmentation maps of images~\cite{NextSegmPredICCV17, Jin_2017_NIPS, Nabavi2018FutureSS, bhattacharyya2018bayesian} or even semantic instance segmentation maps~\cite{Luc_2018_ECCV}. A more difficult problem of future frame prediction has also been explored in~\cite{Ranzato2014VideoM, pmlr-v37-srivastava15, Mathieu2016DeepMV}. Some works have also addressed the task of generating sentences for describing future frames or upcoming steps in recipes~\cite{Sener2019ZeroShotAF}. 

%-------------------------------------------------------------------------

\section{Proposed Approach}
We introduce a multi-modal temporal convolutional network for the task of action anticipation. We start by defining the task of action anticipation in Section~\ref{sec:ant_task}. Then, we discuss the video processing procedure in Section~\ref{sec:video_proc}. Finally, we introduce our uni-modal anticipation branch in Section~\ref{sec:unimodal} and discuss the multi-modal fusion strategy in Section~\ref{sec:fusion}.

\subsection{The Anticipation Task}
\label{sec:ant_task}

We adopt the problem definition of action anticipation from \cite{Damen2018EPICKITCHENS}.
Let $T_a$ be the \textit{anticipation time}, \ie how many seconds in advance an action is predicted, and $T_o$ be the \textit{observation time}, \ie the length of the observed video segment that precedes the action of interest. For a given action video segment $A = [t_{s}, t_{e}]$, let $t_{s}$ and $t_{e}$ denote times of action start and end respectively. Then, the goal of the action anticipation task is to predict the action label of $A$ by observing a video segment of length $T_o$ preceding the action start time $t_s$ by the anticipation time of $T_a$ seconds, that is $[t_{s}-(T_a + T_o), t_{s} - T_a]$. The action anticipation task is illustrated in Figure~\ref{fig:intro}.

In our work, the anticipation time $T_a$ is one second. \Ie an action is anticipated one second before its occurrence. The observation time $T_o$ is set to 5.25 seconds. We will show in the experiments the effect of varying the length of the observed video segment on the performance. %accuracy of the predictions.

\subsection{Video Processing}
\label{sec:video_proc}
We process the observed video segments in the same way as proposed in~\cite{furnari2019rulstm}. 
Let the observed video segment be denoted by $V$. As mentioned previously, we set the length of $V$ to $5.25$ seconds (\ie $T_o = 5.25$). During processing, we break $V$ down into snippets that have a duration of $\alpha = 0.25$ seconds, which results in a total of $N = 21$ snippets $\{V_1, V_2, \dots, V_N\}$. Since the anticipation time is equal to 1.0 second, the resulting video snippets are located at $\{6.0, 5.75, \dots, 1.0\}$ seconds before the action of interest, where the location is defined by the last frame of the snippet.
Each snippet contains several video frames $V_i = \{I^{1}_i, I^{2}_i, \dots, I^{m}_i\}$, where the exact number of frames $m$ depends on the frame rate of the videos. From each snippet, we sample one or several frames, depending on the modality, which are then used to extract features.

In our model, we use three types of modalities: RGB, optical flow and object features. For extracting the RGB features, each snippet is represented by its last frame $I^{m}_i$. For optical flow, the last 5 frames $\{I^{m - 4}_i, \dots, I^{m}_i\}$ of horizontal and vertical flow are used. To get object features, as for the RGB features, the last frame of each video snippet is used, that is $I^{m}_i$. Having collected the frames, we extract RGB and optical flow features using TBN~\cite{kazakos2019TBN} pre-trained for action recognition, while for object features we use the representation proposed by \cite{furnari2019rulstm}. We elaborate on the feature extraction procedure in the implementation details.

\subsection{Uni-modal Branch}
\label{sec:unimodal}

\begin{figure*}
    \begin{center}
    \includegraphics[scale=0.35]{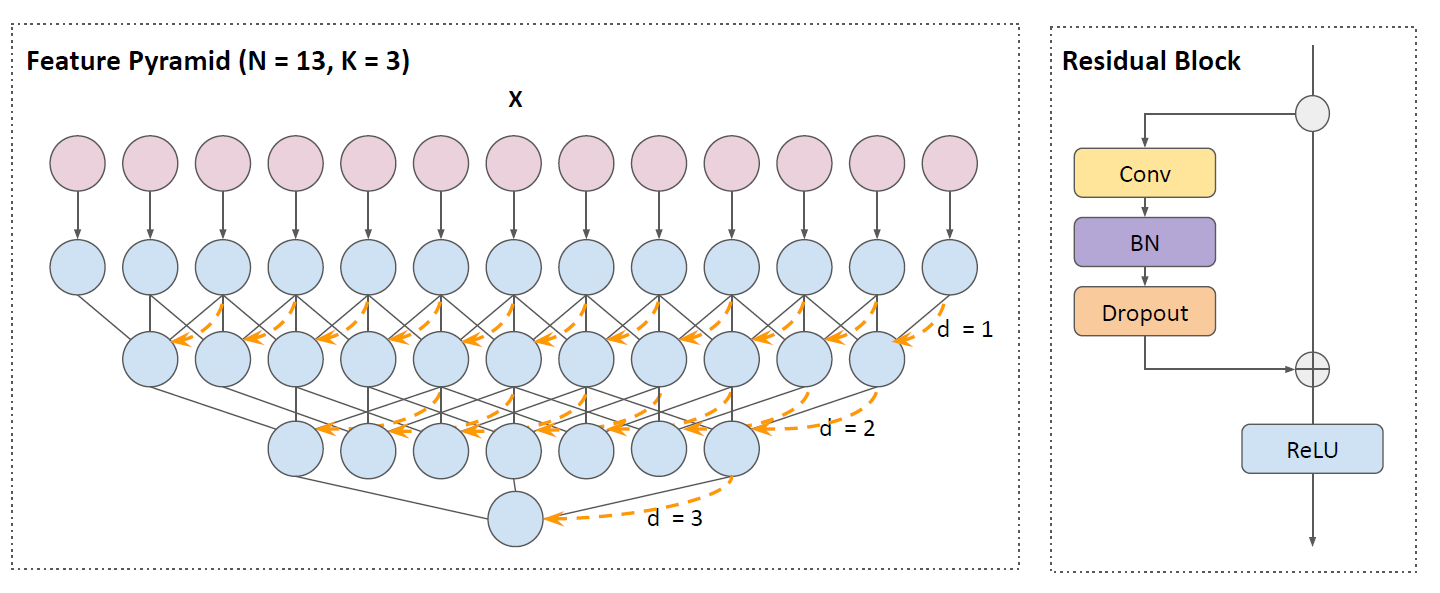}
    \caption{Architectural elements of our uni-modal branch. \textit{(Left)} Our branch consists of a set of dilated temporal convolutions (with kernel size $K$) that process the input sequence (of length $N$) iteratively allowing our model to learn hierarchical feature representations. The increasing dilation factor allows to increase the receptive field of the model without increasing its depth. \textit{(Right)} Overview of the convolutional block with the residual connection. Since the length of the input and output sequences for the residual blocks differ, we only consider the most recent elements of the input sequence for residual connections.}
    \label{fig:model}
    \end{center}
\end{figure*}

Our proposed branch for uni-modal action anticipation is inspired by the temporal convolutional network (TCN) proposed in~\cite{Lea_2017_CVPR}.
Similar to~\cite{Lea_2017_CVPR}, the uni-modal branch stacks several layers of temporal convolutional residual blocks. %, where each block consists of a batch-normalized dilated temporal convolutional layer followed by dropout and ReLU activation. 
An overview of the uni-modal branch is illustrated in Figure~\ref{fig:model}.

The first layer of the proposed branch is a one dimensional convolution with kernel size one, that adjusts the number of channels in the features of the input sequence to match the number of the features maps in the network. After that, the embedded sequence is processed by the residual block layers, that contain dilated one dimensional convolutional filters. These blocks are applied to the input hierarchically, meaning that each layer processes the output of the previous one.

All blocks follow the structure depicted in Figure~\ref{fig:model}. It consists of a temporal convolution with kernel size $K$ and $C$ convolutional filters. 
In our work, we set the kernel size $K = 3$ and the number of filters $C = 1024$. Depending on the layer number $l$, the convolutional filters within the residual blocks have different dilation factors. We increase the dilation factor for the convolutional kernels linearly with the number of layers (\ie 1, 2, 3, 4). 

%Like in \cite{Lea_2017_CVPR}, depending on the layer number $l$, the convolutional filters within the blocks have different dilation factors. While TCN network uses filters with dilation doubled at each level (i.e. 1, 2, 4, 8), we increase the dilation factor for the convolutional kernels linearly (i.e. 1, 2, 3, 4). 

Also, within each block, we normalize the output of the convolution using batch normalization~\cite{Szegedy_2015_CVPR} and apply dropout~\cite{Tompson_CVPR_2015, JMLR:v15:srivastava14a} to avoid overfitting. We also apply dropout to the input of the first dimension-adjustment layer. Similar to \cite{bai2018empirical}, we used a spatial dropout, that is at each training step a whole channel across all time steps is zeroed out.
To facilitate the gradient flow, we further introduce residual connections into the blocks that are followed by ReLU non-linearity. Since the temporal length of the output of the convolutional layer is less than the input sequence, we only use the most recent elements of the input sequence for the residual connections.
%Since after the application of temporal convolution the length of the output sequence is reduced along the temporal dimension, for residual connections we use as many most recent elements of the input sequence as required to match the length of the resulting output sequence along the time dimension.
Formally, the output at the $l^{th}$ level is computed as follows:
\begin{align*}
    \tilde{Z}_l &= BN(W_l * Z_{l - 1} + b_l) \\
    %\hat{Z}_l &= BN(\hat{Z}_l) \\
    \hat{Z}_l &= Dropout(\tilde{Z}_l) \\
    Z_l &= ReLU(\hat{Z}_l + Z_{l - 1}^{[N_{l - 1} - N_l + 1, \dots, N_{l - 1}]}),
\end{align*}

\begin{comment}
\begin{align*}
    \tilde{Z}_l &= BN(W_l * Z_{l - 1} + b_l) \\
    %\hat{Z}_l &= BN(\hat{Z}_l) \\
    \hat{Z}_l &= Dropout(\tilde{Z}_l) \\
    Z_l &= ReLU(\hat{Z}_l + Z_{l - 1}),
\end{align*}
\end{comment}

where $Z_l$ is the output of layer $l$, $*$ denotes the convolution operator with convolutional filters parametrized by a weight matrix \mbox{$W_l \in \mathbb{R}^{K \times C \times C}$} and a bias vector \mbox{$b_l \in \mathbb{R}^{C}$}, $Z_{l - 1}^{[N_{l - 1} - N_l + 1, \dots, N_{l - 1}]}$ is the sub-sequence of the most recent $N_l$ elements of the sequence $Z_{l - 1}$, where $N_l$ denotes length of the sequence at level $l$. 

Overall, the branch contains $L$ layers of the previously discussed blocks. 
We set $L = 4$ for our model. 
Given an input sequence, we pass it through the hierarchy of the residual block layers until the whole sequence is summarized in a single feature vector $F$ at the bottom of the pyramid. 
%In the case of the input sequence with length $N = 21$, it has four layers ($L = 4$). The input sequence goes through the hierarchy of temporal convolution layers until the whole sequence is summarized in a single feature vector, $F$, at the bottom of the pyramid. 
Based on the final feature vector $F$, we perform action anticipation using a fully-connected layer. In addition to the action classification layer, similar to~\cite{sener-2020-temporal}, we also add two more layers for verb and noun classification that solve the auxiliary classification tasks to help anticipation. 

\subsection{Multi-Modal Fusion}
\label{sec:fusion}

\begin{figure}[tb]
    \centering
    \includegraphics[width=\columnwidth]{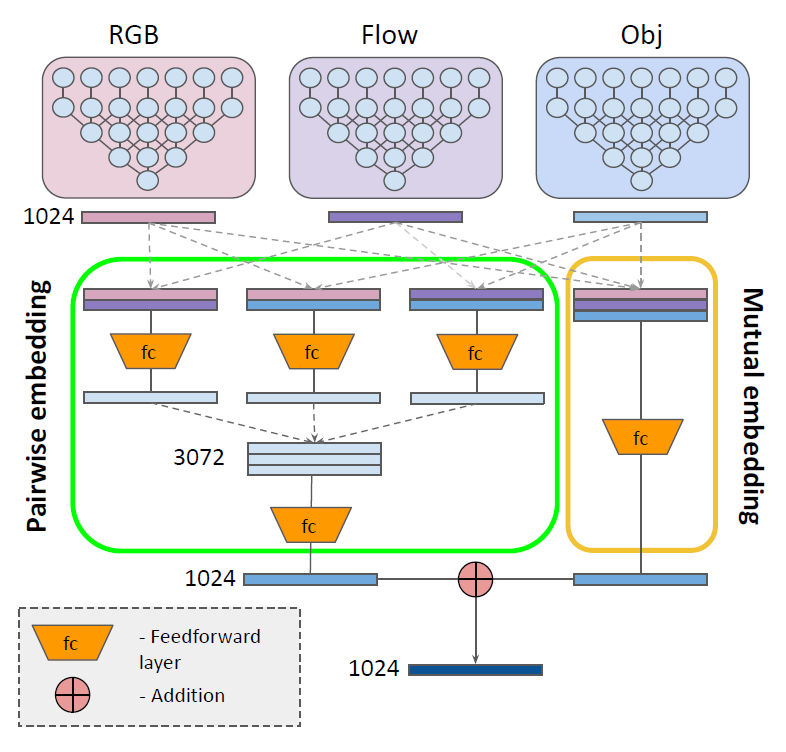}
    \caption{Overview of the multi-modal fusion strategy. Given the output of the individual uni-modal branches, we construct pairwise and mutual embeddings. Finally, these two embeddings are combined and passed to the final classification layers.}
    \label{fig:fusion}
\end{figure}

Our approach fuses three modalities for anticipating the future action: RGB, flow, and object modalities. Figure~\ref{fig:fusion} illustrates the proposed multi-modal fusion strategy.
%The overview of the proposed fusion strategy is illustrated in Figure~\ref{fig:fusion}.
We fuse the branches by constructing a mutual multi-modal feature and use it for performing the final future prediction. To do so, we at first separately pre-train modality-specific branches for action anticipation. Then, 
we extract features $F^{mod}$ from each modality by taking the output of the last convolutional block from the pre-trained branches.
%with the help of the pre-trained networks, whose parameters remain fixed after the networks have been trained, we extract corresponding final features $F^{mod}$ from each branch. These features are taken from the ouput of the last convolutional block of the branches.

We construct a cross-branch multi-modal feature by combining pairwise and mutual embeddings of the features computed by individual branches. 
To compute a pairwise embedding, we at first apply corresponding fully-connected layers to the pairwise concatenations of the features from the three branches. After that, these intermediate representations are merged by another feed-forward layer. A mutual embedding is constructed by projecting the concatenation of the features from the three branches with a fully-connected layer. The output dimension of both pairwise and mutual embeddings is 1024.
Finally, the two computed embeddings are combined by taking their element-wise sum. Based on the resulting feature, three parallel fully-connected classification layers predict action, verb, and noun, respectively.
We demonstrate the effect of using different fusion strategies in the experimental section.

%-------------------------------------------------------------------------

\section{Experiments}

\subsection{Implementation Details}
\subsubsection{Features Extraction}
%We use publicly available RGB frames and pre-computed Optical flow frames for the corresponding datasets \cite{Damen2018EPICKITCHENS, Damen2020RESCALING}.
To extract the RGB and optical flow features, we employ the TBN action recognition network~\cite{kazakos2019TBN}. We use the TBN pre-trained for action recognition, where for pre-training we follow the procedure recommended in~\cite{kazakos2019TBN}. After pre-training, similar to~\cite{furnari2019rulstm}, we take RGB frames and stacks of 5 horizontal and 5 vertical optical flow frames of size $456 \times 256$ and pass them to the network to extract features.
As features, we consider the output produced by the global average pooling layer of the BN-Inception network for the corresponding modality stream, with the resulting representation for each modality containing 1024 channels.

For object features, we use the representation proposed by~\cite{furnari2019rulstm}. They are extracted using a Faster R-CNN~\cite{ren_2015_nips} object detector with ResNet-101 backbone~\cite{he_2016_cvpr}. For all modalities, the corresponding features extraction models are fixed and not fine-tuned for the anticipation task.

\subsubsection{Training Details}
We implemented our model using the Pytorch framework~\cite{pytroch_Neurips_2019}. For optimization we use Stochastic Gradient Descent (SGD) with momentum equal to 0.9 and weight decay of $5 \cdot 10^{-4}$. We trained both the uni-modal branches and the fusion layers for 80 epochs. At first, we separately pre-train uni-modal branches for action anticipation. Then, during fusion, we fix the weights of the individual branches, and optimize only the paramters of the fusion layers.

For EPIC-Kitchens-55, we use a batch size of 64 examples. 
The starting learning rate is set to 0.005 for the object branch and 0.0005 
for the RGB and flow branches as well as for the fusion layers. At each epoch we adjust the learning rate $l$ according to the following schedule: \mbox{$(1 - \frac{e}{E}) ^ {0.99}$}, where $e$ denotes the number of the current epoch and $E$ is the total number of epochs.
For the residual blocks within the individual branches we use a dropout ratio of $0.5$. We also apply dropout of $0.3$ to the input sequence, before applying the first layer. Fully-connected classification layers for uni-modal and multi-modal predictions have dropout rates of $0.7$ and $0.8$, respectively.

For EPIC-Kitchens-100, we use a batch size of 128. We use the same starting learning rates for the uni-modal branches, while for the fusion layers we increase the learning rate to 0.00075. The schedule for the update of the learning rate remains the same.
The dropout ratio for the blocks within the individual branches is decreased to $0.3$ since the dataset is larger and thus the model is less prone to overfitting. As previously, we also apply dropout of $0.3$ to the input sequence. Fully-connected classification layers for both uni-modal and multi-modal predictions have the dropout rate of $0.7$.

For reporting results on the test sets, we use models that are pre-trained on both validation and train splits, after optimizing for hyper-parameters on the validation split.

\subsection{Datasets}
We perform our experiments on two large-scale datasets of egocentric videos: EPIC-Kitchens-55~\cite{Damen2018EPICKITCHENS} and EPIC-Kitchens-100~\cite{Damen2020RESCALING}.

\textbf{EPIC-Kitchens-55} contains videos collected by 32 participants, capturing their daily kitchen activities, including cooking, cleaning, doing laundry, etc. The RGB frames and pre-computed optical flow frames are publicly available with the dataset.
In total, there are 55 hours of recordings and 39596 action annotations. In annotations, there are 125 verb and 352 noun classes. For action classes, similar to~\cite{furnari2019rulstm}, we considered all unique (verb, noun) pairs that are present in the public training set. This amounts to the total of 2513 action classes.

For evaluation purposes, the authors of the dataset defined two test splits: \textit{seen} and \textit{unseen} kitchens. The seen split (S1) contains videos from the kitchens present both in training and test sets, while the unseen test split (S2) contains videos only from such kitchens that have not been observed during training. As a validation set, we use the same subset of training videos as proposed by~\cite{furnari2019rulstm}, who created training and validation sets from the publicly available training set by randomly choosing 232 and 40 videos for each set, respectively.

\textbf{EPIC-Kitchens-100}~\cite{Damen2020RESCALING} extends upon EPIC-Kitchens-55 dataset, with the total of 100 hours of footage. Videos in EPIC-Kitchens-100 were collected by 37 participants in 45 environments.
It contains 89977 fine-grained action annotations, with 97 verb and 300 noun classes. By using the same principle as for EPIC-Kitchens-55, there are 3806 action classes. 
All videos in the dataset are split into train, validation and test sets with a ratio of approximately 75/10/15. The validation and test splits contain two subsets, on which the results are reported separately: unseen participants and tail classes. The unseen subset contains videos of participants that are not present in the train set. The subset of tail classes for verbs and nouns contains the set of classes that have the fewest instances and that account for 20\% of the instances in the whole dataset. An action class is considered to be a tail class, if it contains either a tail noun or verb. Overall, there are 86/228/3729 verb/noun/action classes.

\subsection{Evaluation Metrics}
For both EPIC-Kitchens-55 and EPIC-Kitchens-100, we evaluate our approach using the official dataset metrics. For EPIC-Kitchens-55, we use top-1 and top-5 verb, noun and action accuracy (the prediction is deemed correct if the ground-truth action falls into the top-1 or top-5 predictions, respectively). For EPIC-Kitchens-100, we report class-mean top-5 recall. 
%Evaluation of the model's performance with respect to top-k criteria allows to take into consideration uncertainty of the future prediction. 

%-------------------------------------------------------------------------

%\section{Results}

% COMPARISON TO STATE-OF-THE-ART EPIC-55%
\begin{table*}
\centering
\begin{tabular}{|c|c|c|c|c|c|c|c|}
    \hline
    & & \multicolumn{3}{c|}{Top-1 Accuracy (\%)} & \multicolumn{3}{c|}{Top-5 Accuracy (\%)}\\ 
    \hline
    & Method & Verb & Noun & Action & Verb & Noun & Action \\
    \hline
    \multirow{8}{*}{\textbf{S1}} 
    & TSN~\cite{Damen2018EPICKITCHENS} & 31.8 & 16.2 & 6.0 & 76.6 & 42.2 & 28.2 \\
    & Miech~\etal~\cite{Miech_2019_CVPR_Workshops} & 30.7 & 16.5 & 9.7 & 76.2 & 42.7 & 25.4 \\
    & RU-LSTM~\cite{furnari2019rulstm} & 33.0 & 22.8 & 14.4 & 79.6 & 50.9 &  33.7 \\
    & ImagineRNN~\cite{Wu_2021} & 35.4 & 22.8 & 14.7 & 79.7 & 52.1 & 34.9 \\
    & Liu~\etal~\cite{Liu_2020_ECCV} & 36.3 & 23.8 & 15.4 & 79.2 & 51.9 & 34.3 \\
    & Ego-OMG~\cite{Dessalene_2021} & 32.2 & 24.9 & 16.0 & 77.4 & 50.2 & 34.5\\
    & Sener~\etal~\cite{sener-2020-temporal} & 37.9 & 24.1 & \textbf{16.6} & 79.7 & 54.0 & \textbf{36.1} \\
    & \textbf{Ours (TSN)} & 36.7 & 22.9 & 14.9 & 79.6 & 51.2 & 33.6  \\
    & \textbf{Ours (TBN)} & 37.2 & 23.7 & 15.4 & 79.5 & 51.9 & 34.4 \\
    \hline 
    \multirow{8}{*}{\textbf{S2}}& TSN \cite{Damen2018EPICKITCHENS} & 25.3 & 10.4 & 2.4 & 68.3 & 29.5 & 6.6 \\
    & Miech~\etal~\cite{Miech_2019_CVPR_Workshops} & 28.4 & 12.4 & 7.2 & 69.9 & 32.2 & 19.3 \\
    & RU-LSTM~\cite{furnari2019rulstm} & 27.0 & 15.2 & 8.2 & 69.6 & 34.4 &  21.1 \\
    & ImagineRNN~\cite{Wu_2021} & 29.3 & 15.5 & 9.3 & 70.7 & 35.8 & 22.2 \\
    & Liu~\etal~\cite{Liu_2020_ECCV} & 29.9 & 16.8 & 9.9 & 71.8 & 38.9 & 23.7 \\
    & Sener~\etal~\cite{sener-2020-temporal} & 29.5 & 16.5 & 10.1 & 70.1 & 37.8 & 23.4 \\
    & Ego-OMG~\cite{Dessalene_2021} & 27.4 & 17.7 & \textbf{11.8} & 68.6 & 37.9 & \textbf{23.8} \\
    & \textbf{Ours (TSN)} & 29.3 & 15.2 & 8.9 & 71.2 & 36.8 & 21.0   \\
    & \textbf{Ours (TBN)} & 30.7 & 14.9 & 8.9 & 72.0 & 36.7 & 21.7 \\
    \hline 
    \end{tabular}
    \vspace{1mm}
    \caption{Results for action anticipation on the EPIC-Kitchens-55 seen (S1) and unseen (S2) test splits at anticipation time $T_a = 1$ second.}
    \label{tab:res_ek_55}
\end{table*}

\begin{table}[tb]
    \centering
    \begin{tabular}{|c|c|c|}
         \hline
          & \multicolumn{2}{c|}{Time (sec)} \\%{Average speed per epoch (sec)} \\
         \hline
         Method & Training & Inference \\
         \hline
         RU-LSTM~\cite{furnari2019rulstm} & 27.7 & $3.8 \cdot 10^{-4}$\\
         \hline
         \textbf{Ours} & 12.4 & $2.0 \cdot 10^{-4}$  \\
         \hline
    \end{tabular}
    \vspace{1mm}
    \caption{Average training time per epoch and inference time on EPIC-Kitchens-55. `Training' represents training time on the training split. `Inference' represents average inference time per sample on the validation split.}
    \label{tab:speed}
\end{table}

% COMPARISON TO BASELINE on EPIC-100 %
\begin{table*}
\centering
\begin{tabular}{|c|c|c|c|c|c|c|c|c|c|}
    \hline
    \multicolumn{10}{|c|}{Mean Top-5 Recall} \\ 
    \hline
    & \multicolumn{3}{c|}{Overall (\%)} & \multicolumn{3}{c|}{Unseen (\%)} & \multicolumn{3}{c|}{Tail (\%)}\\ 
    \hline
    Method & Verb & Noun & Act. & Verb & Noun & Act. & Verb & Noun & Act. \\
    \hline
    Random & 6.2 & 2.3 & 0.1 & 8.1 & 3.3 & 0.3 & 1.9 & 0.7 & 0.03\\ 
    RU-LSTM~\cite{furnari2019rulstm} & 25.3 & 26.7 & \textbf{11.2} & 19.4 & 26.9 & 9.7 & 17.6 & 15.9 & \textbf{7.9} \\
    \textbf{Ours (TSN)} & 20.4 & 26.6 & 10.9 & 17.9 & 26.9 & 11.1 & 11.7 & 15.2 & 7.0 \\
    \textbf{Ours (TBN)} & 21.5 & 26.8 & 11.0 & 20.8 & 28.3 & \textbf{12.2} & 13.2 & 15.4 & 7.2\\  
    \hline
    \end{tabular}
    \vspace{1mm}
    \caption{Results for action anticipation at anticipation time $T_a = 1$ on the EPIC-Kitchens-100 test set.}
    \label{tab:res_ek_100}
\end{table*}

\subsection{Anticipation Results on EPIC-Kitchens-55}
We compare our proposed model to the state-of-the-art methods on the test splits of the EPIC-Kitchens-55 dataset in Table~\ref{tab:res_ek_55}. 
In our model, the RGB and optical flow features are extracted using TBN. 
%In addition to presenting evaluation results for our model trained with RGB and optical flow features extracted with TBN, 
For a fair comparison with the methods that use TSN features proposed by~\cite{furnari2019rulstm}, namely~\cite{furnari2019rulstm, sener-2020-temporal, Wu_2021}, we also trained a separate model for which we used appearance and motion features provided by the authors. As one can see, both models perform similarly on the unseen test split, with the top-1 action accuracy of 8.9\%, while on the seen test split, the model trained with the TBN features outperforms the model trained with the TSN features by a margin of 0.5\%. 
Comparing to other methods, on the seen test split, both models trained with TSN and TBN features outperform LSTM-based methods in top-1 action accuracy: RU-LSTM by 0.5\% and 1.0\% accordingly and ImagineRNN~\cite{Wu_2021} by 0.2\% and 0.7\%. On the unseen test split, we outperform RU-LSTM by 0.7\%, while ImagineRNN performs better by 0.4\%. 

Apart from achieving similar or better accuracy, our proposed model is also more efficient during training and inference stages. In Table~\ref{tab:speed} we compare average training time per epoch and time required for inference on the validation set for our RGB branch and that of RU-LSTM. For measuring training and inference times of the RU-LSTM network, we used the official code made publicly available by the authors.
Since RU-LSTM makes predictions at several time steps, to ensure a fair comparison, we performed an experiment where we modified the RU-LSTM training procedure to make predictions at the anticipation time of one second only. This however, resulted in a decrease of both Top-1 and Top-5 accuracy of the model. Therefore, we compare the training time of our method to the original setup of RU-LSTM. Then, for measuring the time required for inference, we considered predictions made by RU-LSTM only for the anticipation time of one second, without performing computations for different anticipation times.
To further minimize the effect of factors not related to the direct effectiveness of the models, for both methods we used TSN features, identical training and validation sets, as well as the same GPU and data loading procedure. We conducted the experiments using an Nvidia Titan Xp GPU.

As shown in Table~\ref{tab:speed}, our method is more than two times faster than RU-LSTM during training and almost two times faster during inference stage. Apart from shorter per-epoch training time, our method also does not use an additional teacher-forcing pre-training stage, as well as the total number of epochs required for convergence is 80 compared to 100 for the RU-LSTM. So, all things considered, our proposed branch can be trained approximately five times faster.
ImagineRNN builds upon the RU-LSTM baseline by extending it with contrastive learning, while the underlying architecture, along with training and inference procedures are identical to those of RU-LSTM, except for the absence of the additional pre-training stage.
Therefore, based on the measurements made for the RU-LSTM, we can expect that both methods require a similar amount of time per-epoch for training and inference. Thus, we can also expect our method to  be faster than ImagineRNN.
Finally, our model is also more effective than LSTM-based models in terms of memory-usage. Our branch needs less than $60\%$ of the memory requirement for the one of the RU-LSTM. Higher memory efficiency of convolution-based networks over RNNs has also been discussed in~\cite{bai2018empirical}.

Concerning the other methods, our model performs on par with Liu~\etal~\cite{Liu_2020_ECCV} on the seen test split, while Ego-OMG and Sener~\etal~\cite{sener-2020-temporal} outperform our approach in top-1 action accuracy by 0.6\% and 1.2\% respectively. On the unseen test split, Liu~\etal~\cite{Liu_2020_ECCV}, Ego-OMG and Sener~\etal~\cite{sener-2020-temporal} perform better than our model by 1.0\%, 2.9\% and 1.2\% respectively.
Notice, however, that both Ego-OMG\cite{Dessalene2020EgocentricOM} and Liu~\etal~\cite{Liu_2020_ECCV} use additional annotations to train their proposed approaches. Ego-OMG uses additional supervision in the form of progression time of directed hand movements, as well as ground truth segmentation masks of interaction objects, while Liu~\etal~\cite{Liu_2020_ECCV} use additional annotations for interaction hotspots and hand trajectories.

\subsection{Anticipation Results on EPIC-Kitchens-100}
We compare our proposed model to the baseline methods on the test set of the EPIC-Kitchens-100 dataset in Table~\ref{tab:res_ek_100}. Since EPIC-Kitchens-100 has been introduced only recently, for many of the previously mentioned methods no evaluation results are available. Therefore, we compare our method to the officially reported baselines. 

As the table shows, our approach performs on par with RU-LSTM on both overall and tail-class splits. The performance is also consistent using different types of features and our model works well with both TBN and TSN features. 
%Using TBN features, the difference in mean top-5 action recall is 0.2\% and 0.5\% respectively, while for model with TSN features - 0.3\% and 0.7\%. 
Furthermore, our approach shows better generalization behavior as indicated by the results on the unseen environments. Our approach outperforms RU-LSTM in the mean top-5 action recall on the unseen split by 2.5\% and 1.4\% using TBN and TSN features, respectively.
%On the unseen split, however, both of our methods perform better than RU-LSTM by a margin of 2.5\% and 1.4\% for TBN and TSN respectively.
%So we can see that our method shows better generalization behavior to the unseen environments.

%-------------------------------------------------------------------------
\subsection{Ablation Study}
In this section, we provide a set of ablation experiments to analyze the different components of our approach. 
%We verify the validity of our proposed model's elements and parameters via a series of ablation studies. 
As our main motivation is to develop a model that has a good trade-off between accuracy and efficiency for the task of action anticipation, we verify how much past is really necessary for the network to achieve a good accuracy. Additionally, we also study different fusion strategies for the uni-modal branches.
For the ablation experiments, we report results on the validation set of the EPIC-Kitchens-55 dataset using TBN features. As previously, we use the validation set constructed by~\cite{furnari2019rulstm}.

\subsubsection{Observation length}
We report the top-1 action accuracy of our RGB branch trained on observation intervals of different lengths in Table~\ref{tab:num_segments}.
We vary the observation length starting from 0.75 seconds up to 7.75 seconds. 
%For simplicity, we chose the observation times in such a way, that the resulting input sequence is summarized into a single representation vector by a fixed number of layers. 
%In general, it does not necessarily have to be the case and the final feature can for example be created by pooling the output of the $L$ temporal convolutional layers along the time dimension.
As shown in the table, the accuracy of the predictions increases with the length of the observation interval until it saturates at 5.25 seconds. Similar findings have been reported in~\cite{sener-2020-temporal}.
Based on this observation, we fix the length of the observed interval to 5.25 seconds which corresponds to an input sequence of 21 snippets.

% NUMBER OF SEGMENTS COMPARISON ON EPIC-55 %
\begin{table}[]
    \centering
    \resizebox{\columnwidth}{!}{%
    \begin{tabular}{|c|c|c|c|c|}
    \hline
    No. of Snippets & Obs. Time (sec) & Top-1 Act. Acc (\%) \\
    \hline
    3  & 0.75 & 11.1\\ 
    \hline
    7  & 1.75 & 11.5\\
    \hline
    13 & 3.25 & 11.8 \\
    \hline
    21 & 5.25 & \textbf{12.4} \\
    \hline
    31 & 7.75 & \textbf{12.4} \\
    \hline
    \end{tabular}
    }
    \vspace{1mm}
    \caption{Effect of the observation length on the prediction accuracy. We report top-1 action accuracy for the RGB branch using TBN features on the validation set of EPIC-Kitchens-55.}
    \label{tab:num_segments}
\end{table}

\subsubsection{Fusion strategy}
We report the top-1 action accuracy of the individual branches and different multi-modal fusion strategies in Table~\ref{tab:fusion}.
Among the uni-modal branches, the appearance branch has the highest top-1 action accuracy, whereas the flow branch has the lowest. All fusion schemes improve over the performance of uni-modal branches. In total, we experimented with five different fusion methods:
\begin{itemize}
    \item \textbf{Late fusion}: Averaging predictions made by individual branches.
    \item \textbf{Attention fusion}: Learning weights for predictions of the individual branches based on the final features from the three branches: $\{F^{RGB}, F^{Flow}, F^{Object}\}$.
    \item \textbf{Mutual fusion}: Applying only the mutual fusion path from the proposed fusion scheme (see Figure~\ref{fig:fusion}).
    \item \textbf{Pairwise fusion}: Applying only the pairwise fusion path from the proposed fusion scheme (see Figure~\ref{fig:fusion}).
    \item \textbf{Mutual + Pairwise}: Applying both pairwise and mutual fusion paths and merging them with element-wise addition (see Figure~\ref{fig:fusion}).
\end{itemize}
We observed that merging predictions of individual branches either via late fusion (top-1 action accuracy 14.1\%) or attention fusion (top-1 action accuracy 14.3\%) achieves lower results than merging individual features of the uni-modal branches and making a cross-branch prediction based on the constructed representations. Furthermore, learning both pairwise and mutual embeddings of the uni-modal features and combining them via the element-wise addition is better than making the final prediction based only either on the pairwise or mutual embeddings. 

% FUSION STRATEGY COMPARISON ON EPIC-55 %
\begin{table}[]
    \centering
    \begin{tabular}{|c|c|}
         \hline
         Fusion & Top-1 Act. Acc (\%) \\
         \hline
         \hline
         RGB & 12.4 \\
         \hline
         Flow & 8.9 \\
         \hline
         Obj & 10.9 \\
         \hline
         \hline
         Late fusion & 14.1  \\
         \hline
         Attention &  14.3 \\
         \hline
         Mutual feature fusion &  14.7\\
         \hline
         Pairwise feature fusion & 14.6 \\
         \hline
         Mutual + Pairwise feature fusion & \textbf{14.9} \\
         \hline
    \end{tabular}
    \vspace{1mm}
    \caption{Comparison of different multi-modal fusion methods on the EPIC-Kitchens-55 validation set.}
    \label{tab:fusion}
\end{table}

%-------------------------------------------------------------------------

\section{Conclusion}
In this work, we proposed a multi-modal architecture based on temporal convolutional layers for the short-term action anticipation task. Instead of relying on recurrent layers for temporal modelling, we use a stack of temporal convolutional layers, which allows our approach to perform anticipation faster. We further proposed a multi-modal fusion strategy that combines both mutual and pairwise interactions between the different branches. Results on two large-scale datasets of egocentric videos, EPIC-Kitchens-55 and EPIC-Kitchens-100, show that our approach achieves performance comparable to the state-of-the-art approaches while being at least two times faster and more efficient compared to RNN-based approaches.

%-------------------------------------------------------------------------

\section{Acknowledgements}
This work has been funded by the Deutsche 
Forschungsgemeinschaft (DFG, German Research Foundation) – GA 1927/4-2 
(FOR 2535 AnticipatingHuman Behavior) and the ERC Starting Grant ARCA 
(677650).

{\small
\bibliographystyle{ieee_fullname}
\bibliography{egbib}
}

\end{document}